\documentclass[letterpaper, 10pt, conference]{ieeeconf}

\usepackage[left=54pt,top=50pt,right=54pt,bottom=57pt]{geometry}
\usepackage{times}
\usepackage{tabularx}
\usepackage{subcaption}
\usepackage{multirow,multicol, array}
\usepackage[font={small}]{caption}   
\usepackage{graphicx}
\usepackage{wrapfig}

\usepackage{algorithm}
\usepackage[noend]{algpseudocode}

\let\labelindent\relax
\usepackage{enumitem}

\usepackage{amsmath,amsthm,amssymb,amsfonts}
\usepackage{textcomp}
\usepackage{acronym}
\usepackage{balance}
\usepackage{mdwmath}
\usepackage{bm}
\usepackage{mdwtab}
\usepackage{array}
\usepackage[usenames,dvipsnames]{color}
\usepackage{eqparbox}
\usepackage{cite}

\usepackage{color}
\usepackage{psfrag}
\usepackage{epsfig}
\usepackage{url}

\usepackage{epstopdf}
\usepackage{booktabs}
\usepackage{blindtext}

\usepackage{soul}

\usepackage{comment}

\usepackage[colorinlistoftodos,prependcaption,textsize=small]{todonotes}

\renewcommand{\emph}{\textit}

\newtheorem*{lemma*}{Lemma}

\newtheorem*{problem*}{Problem}

\makeatletter
\def\maketag@@@#1{\hbox{\m@th\normalfont\normalsize#1}}
\makeatother

\IEEEoverridecommandlockouts
\graphicspath{{images/}}

\usepackage{siunitx}

\begin{document}

\title{\LARGE \bf
A Portable Agricultural Robot for Continuous Apparent Soil Electrical Conductivity Measurements to Improve Irrigation Practices
}

\author{Merrick Campbell,$^{1}$ Keran Ye,$^{1}$ Elia Scudiero,$^{2}$ and Konstantinos Karydis$^{1}$
\thanks{$^{1}$M. Campbell, K. Ye, and K. Karydis are with Dept. of Electrical and Computer Eng., Univ. of California, Riverside, 900 University Avenue, Riverside, CA 92521, USA.
{\tt\footnotesize\{mcamp077, kye007, karydis\}@ucr.edu}.}%
\thanks{$^{2}$E. Scudiero is with Dept. of Environmental Sciences, Univ. of California, Riverside, 450 West Big Springs Road, Riverside, CA 92507, USA.
{\tt\footnotesize elia.scudiero@ucr.edu}. }
\thanks{
We gratefully acknowledge the support of USDA-NIFA under grant \# 2021-67022-33453, a UC MRPI Award, and a Frank G. and Janice B. Delfino Agricultural Technology Research Initiative Seed Award. 
Any opinions, findings, and conclusions or recommendations expressed in this material are those of the authors and do not necessarily reflect the views of the funding agencies.
}}

\maketitle
\thispagestyle{empty}
\pagestyle{empty}

\begin{abstract}

Near-ground sensing data, such as geospatial measurements of soil apparent electrical conductivity (ECa), are used in precision agriculture to improve farming practices and increase crop yield. Near-ground sensors provide valuable information, yet, the process of collecting, assessing, and interpreting measurements requires significant human labor. Automating parts of this process via the use of mobile robots can help decrease labor burden, and 
increase the accuracy and frequency of data collections, and overall increase the adoption and use of ECa measurement technology. This paper introduces a roboticized means to autonomously perform  geospatial ECa measurements and map soil moisture content in micro-irrigated orchard systems. We retrofit a small wheeled mobile robot with a small electromagnetic induction sensor by studying and taking into consideration the effect of the robot body to the sensor's readings, and develop a software stack to enable autonomous logging of geo-referenced measurements. The proposed roboticized ECa measurement method is evaluated by mapping a 50m x 30m field against the baseline of human-conducted measurements obtained by walking the sensor in the same field and following the same path. Experimental testing reveals that our approach yields roboticized measurements comparable to human-conducted ones, despite the robot's small form factor.

\end{abstract}

\section{Introduction}


Precision agriculture is an increasingly adopted farming practice that aims to administer agronomic inputs (e.g., irrigation, fertilizers, pesticides) when, where, and in the amount needed. To inform such management, networks of ground and remote sensors are used to accurately characterize soil-plant-environment processes~\cite{ZHANG2002113}. Precision irrigation, specifically, can increase grower revenue and decrease the environmental footprint of agriculture by applying water directly where and when required from the ideal source~\cite{ZHANG2002113,vellidis2016dynamic}. Accurate estimates of water available to plant roots throughout the soil profile can be obtained with soil sensors~\cite{https://doi.org/10.1029/WR016i003p00574,CALAMITA2015316}. However, only 12\% of growers in the USA use root-zone soil measurements to trigger and budget irrigation events~\cite{VELLIDIS2016249, USDAsurvey}. The growers that utilize these measurements rely on an expensive network of 
decentralized sensors. These sensors generally measure only a few points across large swathes of land and provide an incomplete picture of irrigation practices. Such lean sampling fails to capture soil spatial variability, which is a key component of plant-water-environment relationships~\cite{nielsen1973spatial}. Knowledge of soil moisture (SM) spatial variability can improve precision irrigation. 

Capturing continuous measurements with a robotic system helps remove several current barriers to entry that prevent the wider adoption of soil moisture measurements. Geospatial electromagnetic induction (EMI) measurements of soil apparent electrical conductivity (ECa) are a reliable proxy for SM spatial variability~\cite{corwin2020field}. Barriers to the use of this technology include the cost of carrying out reliable EMI surveys~\cite{corwin2020field} and calibrating the sensor readings to SM estimations. Concurrent in-situ measurements of soil moisture can calibrate field-scale ECa geospatial surveys to accurately map root-zone soil moisture using physically-based stochastic modeling~\cite{nielsen1973spatial}. Then site-specific ECa-to-SM calibrations are elaborated by experts using concurrent SM data from in-situ SM monitoring stations or from soil cores. 


\begin{figure*}[!t]
\vspace{6pt}
    \centering
    \includegraphics[trim={0cm, 10.5cm, 4.3cm, 0cm},clip,width=0.995\linewidth]{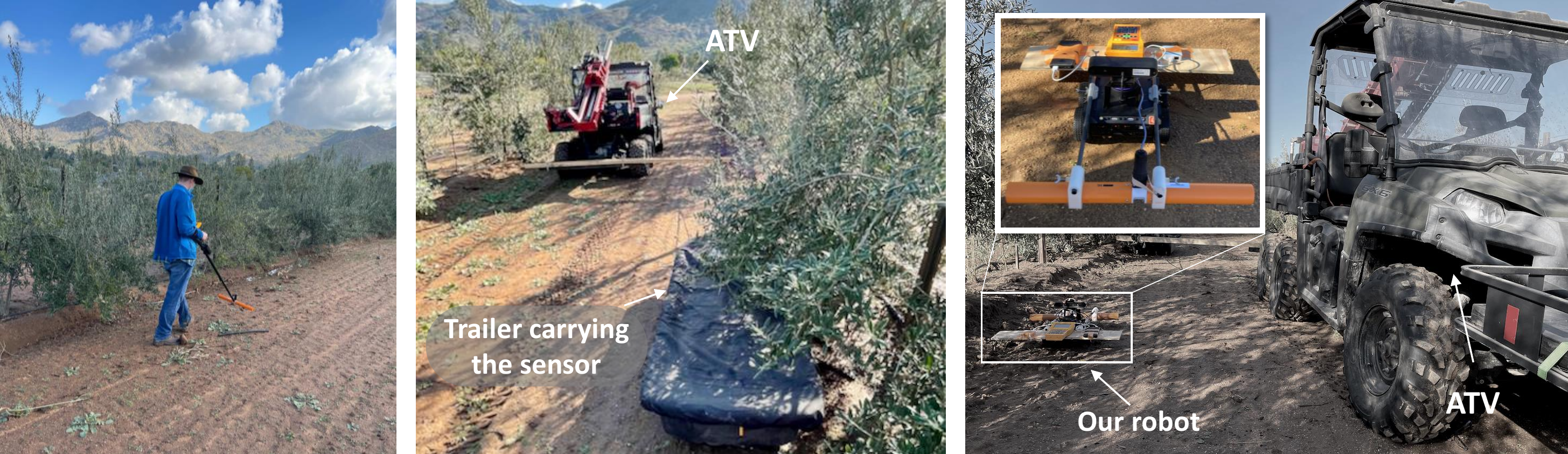}
    \caption{Current and our envisioned soil moisture (SM) survey techniques using electromagnetic induction (EMI) sensors in traditional and precision agriculture. Current methods include manually-collected data obtained by walking the sensor in the field (left panel), and data obtained by a person driving a field vehicle that pulls the sensor secured on a trailer (central panel). Both methods are labor-intensive. In contrast, our proposed method seeks to automate this process via the use of small and portable agricultural robotics (right panel). Herein we consider a ROSbot 2.0 Pro wheeled robot which we retrofit with a CMD-Tiny EMI soil sensor (inset image in right panel) to conduct SM surveys over a $50$\;m $\times$ $30$\;m drip-irrigated olive orchard. Our robot prototype is significantly smaller than the ATV and sensor trailer. Due to its small size, it is able to navigate closer to the drip-lines at the base of the trees and exert more control over the spatial component of apparent Soil Electrical Conductivity (ECa) measurements. The robot contains all necessary equipment to perform the measurement including the EMI sensor, a GNSS receiver, and router to provide a local field network.}
    \label{fig:robo_v_atv}
    \vspace{-12pt}
\end{figure*}


However, integrating an EMI soil sensor with a robotic platform presents several challenges. The EMI sensors use a paired emitter to induce a magnetic field in the soil and receiver to measure such field. The soil ECa is derived from these electromagnetic measurements. Though continuous robotic measurements would benefit irrigation models, this configuration introduces the possibility that metal and electronic components of the robot can interfere with the sensor measurement. The placement and orientation of the sensor on the platform is critical since the mere presence of the robot in proximity to the sensor can interfere with the sensor reading. Sensor placement is also constrained by the robot's size, weight, and power (SWaP) requirements to be both portable and effective at traversing uneven terrain.


In typical applications, SM surveys with EMI sensors are carried out manually either by walking the sensor in the field, or by driving field vehicles with a trailing sensor (see left and center panels of Fig.~\ref{fig:robo_v_atv}, respectively). Both approaches are time-consuming, labor-intensive, and limit broad-scale adoption of this technology for frequent SM mapping. 
Established standard operation procedures~\cite{corwin2020field} recommend that ECa measurements should not be carried out on very dry soils, especially in arid, semi-arid and Mediterranean climates where the space between tree rows of micro-irrigated orchards are generally very dry because irrigation only wets soils very close to the drip emitters~\cite{Skaggs2004ComparisonOH}. 
Depending on soil type and irrigation strategy, moist soil is often found up to $1-1.5$\;m away from drip emitters~\cite{https://doi.org/10.2136/sssaj2009.0341}. In these orchard systems, ECa measurements should thus be carried out close to the trees, along the drip-lines. Previous research on SM sensing robots use larger platforms and very expensive sensor technology such as cosmic-ray sensors~\cite{8711348,KrigingFramework,PHILLIPS201473}. This limits how close the sensors can get to the region of interest and how frequently the surveys can be conducted. Thus, small, portable, and cost-effective SM survey robots may improve ECa measurement accuracy by bringing the EMI sensor closer to the tree roots and increase survey frequency~\cite{https://doi.org/10.1111/sum.12261}.

In this paper we develop an integrated robotic soil sensor platform (right panel and inset in Fig.~\ref{fig:robo_v_atv}) with the aim to investigate whether a compact robot can serve as a reliable and accurate platform for performing continuous ECa measurements as part of a SM survey. 
We use a ROSbot 2.0 pro wheeled mobile robot, and investigate the effect the robot body and carried sensors have on EMI sensor readings. Through a series of experiments we determine an appropriate configuration for mounting the sensor on the robot so as to minimize interference without compromising the robot's mobility and operational envelope. 
We further develop a software stack via the Robot Operating System (ROS) to enable autonomous logging of geo-referenced ECa measurements. The developed platform is evaluated in spatial mapping of a $50$\;m $\times$ $30$\;m drip-irrigated olive orchard, and compared against a spatial map created via manually-taken measurements with the handheld sensor. 
Results demonstrate the efficacy and reliability of our proposed method, thus confirming the viability of the use of small-factor mobile robots in spatio-temporal SM mapping.
Accurate spatio-temporal soil moisture information is key to increasing environmental and economic sustainability of precision irrigation.








\vspace{0pt}
\section{Key Features of the Proposed Integrated Robotic Soil Sensor Platform}



\subsection{Measuring Soil Conductivity}

\begin{figure}[!h]
   \vspace{-3pt}
      \centering
      \includegraphics[trim={0cm 0cm 1.5cm 1.5cm},clip,width=.95\linewidth]{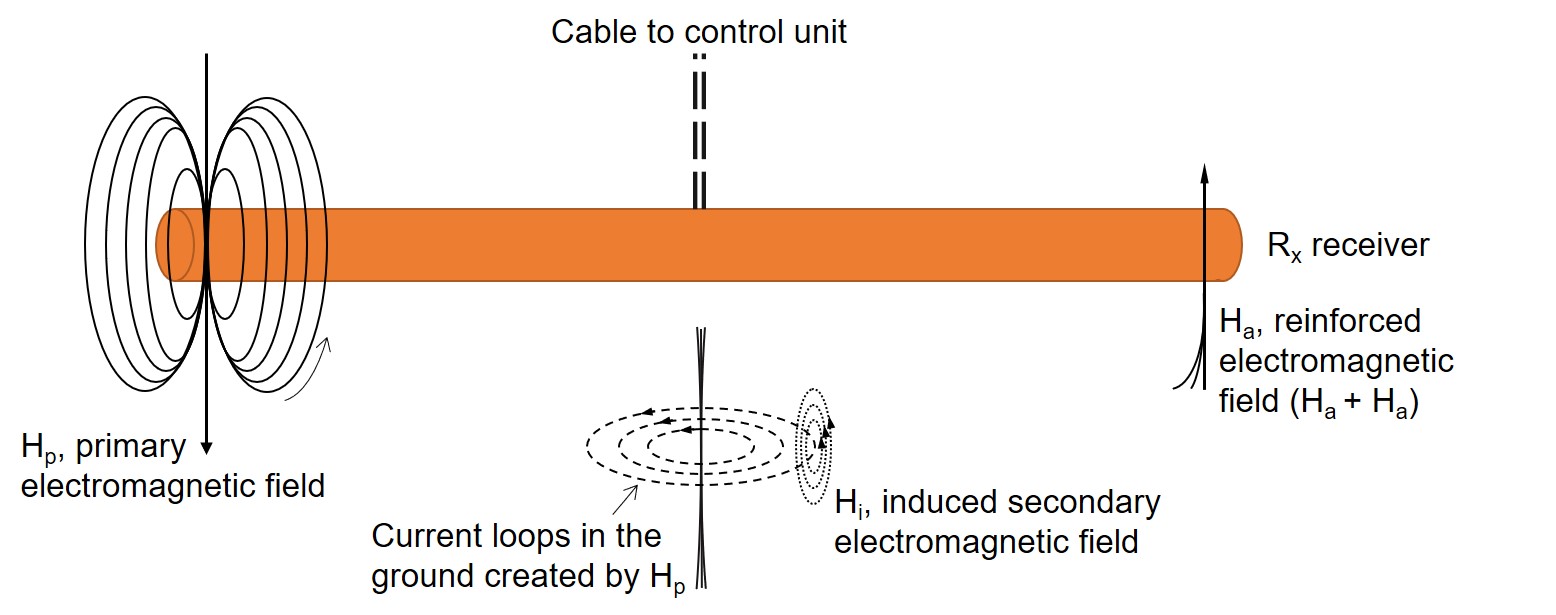}
      \caption{Schematic of the operation of the CMD-Tiny probe. Modified from Lesch et al. \cite{lesch2005apparent}. This sensor can measure apparent conductivity up to 1000\;mS/m with a resolution of 0.1\;mS/m with an accuracy of $\pm 4\%$ at 50\;mS/m.
      }
      \label{sensortheory}
      \vspace{-3pt}
   \end{figure}

Apparent soil electrical conductivity (ECa) measures the bulk conductivity of the soil and the resulting measurement is a complex interaction of salinity, water content, and soil composition~\cite{corwin2020field}. Though the interactions may be complex, soil moisture can be reliably inferred from ECa measurements when ground-truth data are used~\cite{corwin2020field}.  
During operation, the EMI sensor generates a primary electric field that induces eddy currents into moist soil. These currents generate a secondary magnetic field that is then measured by the sensor's receiver~\cite{lesch2005apparent}. Figure~\ref{sensortheory} depicts the sensor's operating principle. Measured ECa values serve as a reliable proxy for soil moisture when the sensor is properly calibrated.

In this work we use a CMD-Tiny (GF Instruments\footnote{GF Instruments: http://www.gfinstruments.cz}) conductivity sensor, which is a small-form sensor representative of many typical sensors on the market. The sensor has diameter $42.5$\;mm and length $500$\;mm and weighs $424$\;g (i.e., only $4$\% of ROSbot's total payload capacity of $10$\;kg). The sensor is paired with a data logger that also provides power to the former. As such, this sensor is a good candidate for integrating with the ROSbot platform.

\subsection{Performing Roboticized ECa Measurements}
\begin{figure}[!h]
\vspace{-3pt}
    \centering
    \includegraphics[width=.9\linewidth]{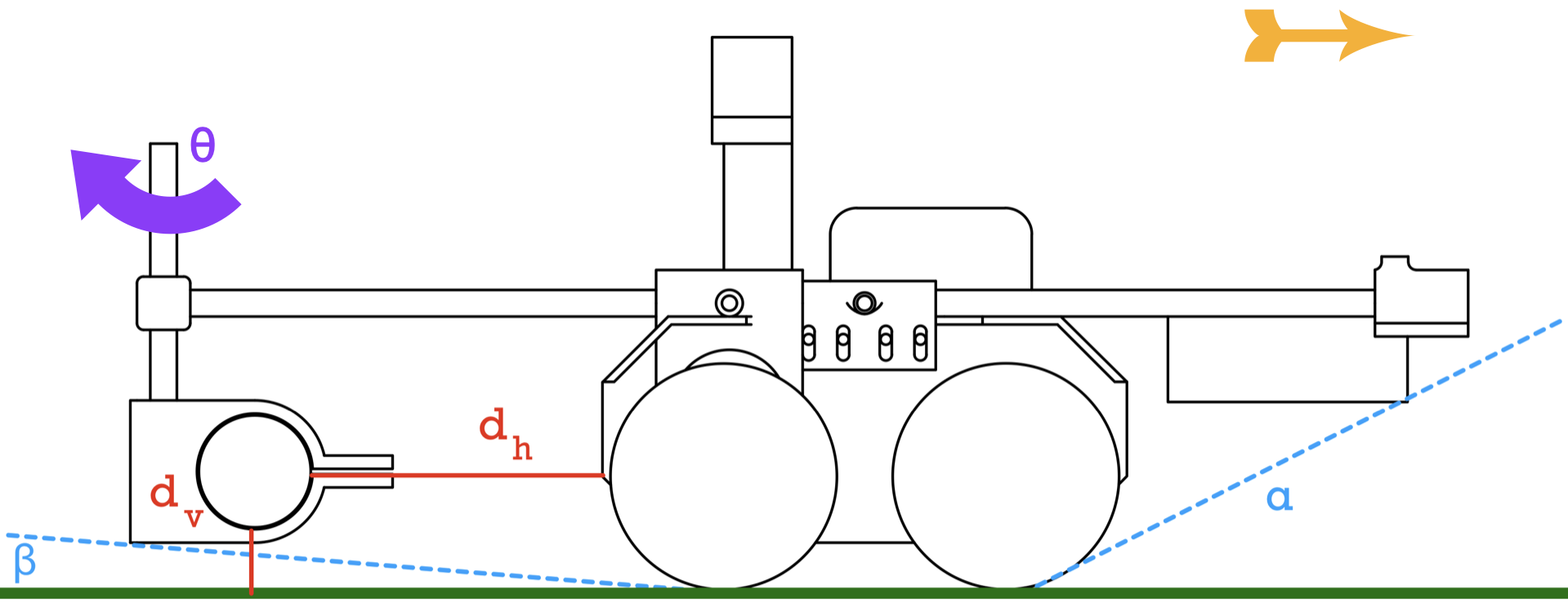}
    \caption{The proposed robotic platform. Design parameters $d_h, d_v, \theta, \alpha$ and $\beta$ are determined in this work. The arrow (shown in yellow) denotes the forward-looking direction of the robot.}
    \label{fig:sketch}
    \vspace{-3pt}
    \end{figure}

The ROSbot 2.0 Pro (Husarion Robotics\footnote{Husarion Robotics: https://husarion.com})~\cite{de2020multi} is adopted as the mobile platform to deploy in the field. This compact and light-weight robot contains the necessary integrated hardware to enable stable maneuverability, remote control, and autonomous navigation. As shown in Fig.~\ref{fig:sketch}, the EMI sensor is orientated with respect to the robot with angle $\theta$, installed with ground clearance $d_v$, and located away from the robot's back with distance $d_h$ to improve the SM measurement accuracy and reduce the robot's interference. The parameters also constrain the approach angle $\alpha$ and the departure angle $\beta$ which could limit the capability of overcoming obstacles in the uneven terrain. With careful design parameter selection (discussed in Section~\ref{sec:tests}), the fully-equipped robot can traverse common field terrains with deviation of $\pm 25$\;mm.

\subsection{System Software Architecture}



Our platform's software architecture is built upon ROS to enable automatic geospatial mapping of ECa measurements and soil moisture content in micro-irrigated orchard systems. The software consists of navigation and ECa modules that are interpreted as several ROS nodes. The navigation module utilizes an external GNSS receiver and ROSbot's odometry information fused from motor encoders and IMU to compute the robot's real-time pose (i.e. position and orientation) with an Extended Kalman Filter~\cite{julier1997new}. Robot control commands are generated from either manual inputs or a waypoint-based trajectory planner. In field testing we used teleoperation; fully autonomous navigation in the field~\cite{barawid2007development,ruckelshausen2009bonirob,hebert2012intelligent,wellington2006generative} by integrating measurements from the onboard LiDAR~\cite{malavazi2018lidar,hiremath2014laser} and RGB-D camera~\cite{endres20133,huang2017visual} activated for online obstacle avoidance falls outside the scope of this work and is part of future work. The ECa module extracts SM information from the EMI sensor and synchronizes the data with spatial coordinates from the navigation module. Geospatial SM data are displayed in real time and simultaneously logged into local CSV files for post-processing with ArcMap 10.8 (ESRI, Redlands, CA).

\section{Experimental Methods and Results}\label{sec:tests}
\subsection{Overview of Experimental Procedures}
Development and testing of our integrated robotic soil sensor platform was conducted over four experimental stages. 
1) First, we performed a series of distinct sampling measurement across diverse fields to validate repeatability and effect of distance, $d_h$, between the sensor and the main robot chassis so as to minimize EMI interference. 
2) Having identified a mapping between distance ($d_h$) and EMI interference to measurements, the second stage involved integration of the sensor onto the robot in a way that the approach ($\alpha$) and departure ($\beta)$ angles are such that the robot can navigate mild uneven terrain. %
3) Following the platform design as per the first two stages, we then performed repeated roboticized ECa measurements over discrete sampling points in the field and compared against manually-conducted measurements at the same spots in order to validate measurement repeatability and accuracy of the roboticized ECa measurements. 
4) Lastly yet importantly, we evaluated the efficacy of our overall roboticized approach in collecting continuous ECa data, and compared its results against manually-collected continuous samples, over the same field.

\subsection{Preliminary Sensor Placement Tests}\label{subsec:prelim}

\textbf{Objective:}
Identify a mapping between sensor and main robot chassis distance and EMI interference, and determine viable sensor placement configurations minimizing the latter.

\begin{figure}[!h]
\vspace{-4pt}
    \centering
    \includegraphics[trim={0cm 1.5cm 0cm 1cm},clip,width=0.75\linewidth]{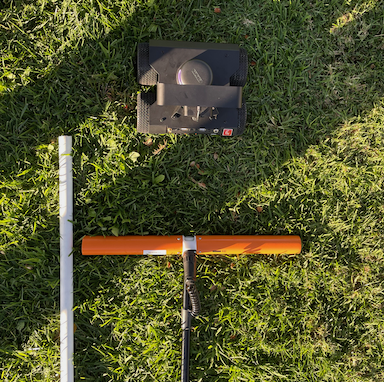}
    \caption{In field tests to identify EMI interference, the sensor (orange cylinder) was operated in a manual configuration to record a series of measurements at multiple distances and fixed orientation (either $0^\circ$ or $90^\circ$) from the robot, over diverse fields. Here we show an instance from testing over irrigated turf at $d_h=25$\;mm and $\theta=0^\circ$. 
    }
    \label{test_sensor_placement}
    \vspace{-4pt}
\end{figure}

\textbf{Setup:}
The robot was placed at different locations of interest, and at different times and days to establish a rich basis, and several discrete ECa measurements were taken at fixed distance intervals from the robot, $d_h=\{25, 30,36, 41, 46, 51, 56, 61\}$\;cm, and at two orientations, $\theta=\{0^\circ,90^\circ\}$. Control measurements without the robot present were also taken at all sampled locations. 
The test environments included grass, bare-soil, and tree roots found at the Agricultural Experimental Station of the University of California, Riverside, and at the USDA-ARS U.S. Salinity Laboratory, also in Riverside, California. Table~\ref{tab:prelim} lists all test cases, 
and Fig.~\ref{test_sensor_placement} depicts an example from testing on irrigated turf. Measurements were performed with the CMD-Tiny sensor in the `high' mode, which is tuned by the manufacturer to provide high-accuracy ECa measurements in $[0,0.7]$\;m soil depth.


\begin{table*}[!h]
\vspace{6pt}
\centering
\caption{Conductivity Measurements (mS/m) at Varying Distances $d_h$ (cm) Over Various Test Configurations}
\label{tab:prelim}
\resizebox{1.75\columnwidth}{!}{
\begin{tabular}{l l l l l l l l l l}
\textbf{Test Configuration ~~~ || ~~~  Distance:} & \multicolumn{1}{l}{\textbf{$\infty$}} & \textbf{25} & \textbf{30} & \textbf{36} & \textbf{41} & \textbf{46} & \textbf{51} & \textbf{56} & \textbf{61} \\
\toprule
\multirow{7}{*}{Citrus Grove, $\theta=0^\circ$} & 26.7                             & 41.8        & 35.4        & 31.9        & 29.5        & 28.8        & 27.9        & 27.5        & 27.2        \\
         & 19.8                             & 30.3        & 25.9        & 24.3        & 22.1        & 21.1        & 20.8        & 20.4        & 20.1        \\
         & 21.4                             & 32.7        & 28.1        & 26.4        & 24.3        & 23.1        & 22.7        & 22.2        & 21.7        \\
         & 18.3                             & 29.9        & 28          & 25.3        & 23.9        & 22.5        & 21.5        & 21.1        & 20.2        \\
        & 18.1                             & 34.3        & 27.9        & 23.6        & 21.2        & 20          & 19.3        & 18.8        & 18.4        \\
        & 22.4                             & 33.5        & 28          & 26.1        & 24.7        & 23.6        & 22.9        & 23          & 22.7        \\
        & 19.5                             & 35.4        & 30.1        & 26.9        & 23.5        & 22.5        & 21.5        & 21.1        & 20.6        \\
\midrule
\midrule
\multirow{7}{*}{Irrigated Turf, $\theta=0^\circ$}      & 30.6                             & 42.7        & 39.1        & 35.3        & 33.5        & 32.4        & 31.8        & 31.3        & 30.8        \\
& 32.6                             & 47.1        & 42.2        & 38.2        & 35.7        & 34.8        & 33.7        & 33.3        & 32.8        \\
& 26.6                             & 41.5        & 36.1        & 32.3        & 29.6        & 28.8        & 27.9        & 27.1        & 26.9        \\
& 28.20                            & 42.90       & 37.80       & 33.50       & 31.60       & 30.40       & 29.70       & 29.20       & 28.90       \\
& 32.20                            & 47.50       & 41.60       & 38.70       & 35.70       & 34.80       & 33.90       & 33.20       & 33.00       \\
& 28.70                            & 39.20       & 35.10       & 33.10       & 31.30       & 30.40       & 29.80       & 29.50       & 28.90       \\
\midrule
\multirow{3}{*}{Irrigated Turf, $\theta=90^\circ$}     & 28.30                            & 31.20       & 30.20       & 29.40       & 29.10       & 28.50       & 28.60       & 28.40       & 28.30       \\
& 32.40                            & 35.50       & 33.90       & 33.20       & 33.00       & 32.80       & 32.80       & 32.70       & 32.60       \\
& 28.80                            & 31.10       & 29.70       & 29.10       & 28.80       & 28.60       & 28.70       & 28.60       & 28.40       \\
\midrule
\midrule
\multirow{3}{*}{Tree roots, $\theta=0^\circ$}      & 14.90                            & 26.30       & 23.20       & 20.10       & 18.30       & 17.00       & 16.30       & 15.80       & 15.30       \\
& 13.9                             & 28.1        & 23.9        & 19.6        & 16.8        & 15.9        & 15.2        & 14.4        & 14.3        \\
& 16.5                             & 31.1        & 25.2        & 21.5        & 19.1        & 18.3        & 17.6        & 17.2        & 16.9        \\
\midrule
\multirow{1}{*}{Tree roots, $\theta=90^\circ$}     & 14.70                            & 18.30       & 16.60       & 15.70       & 15.20       & 14.90       & 14.90       & 14.70       & 14.70       \\
\midrule
\midrule
\multirow{3}{*}{Bare Soil, $\theta=0^\circ$}     & 23.2                             & 39.5        & 33.5        & 28.1        & 26.7        & 24.9        & 24.3        & 23.8        & 23.5        \\
& 22.40                            & 35.00       & 31.30       & 28.10       & 25.90       & 24.80       & 23.90       & 23.40       & 23.10       \\
& 17.20                            & 25.40       & 23.40       & 21.20       & 19.90       & 19.10       & 18.50       & 18.20       & 17.90       \\
\midrule
\multirow{2}{*}{Bare Soil, $\theta=90^\circ$}     & 22.40                            & 25.20       & 23.90       & 23.30       & 22.90       & 22.70       & 22.60       & 22.50       & 22.50       \\
& 18.50                            & 20.90       & 20.10       & 19.60       & 19.20       & 19.00       & 18.90       & 18.90       & 18.90       \\
\bottomrule
\end{tabular}}
\end{table*}

\begin{figure}[!h]
\vspace{6pt}
    \centering
    \includegraphics[width=0.99\linewidth]{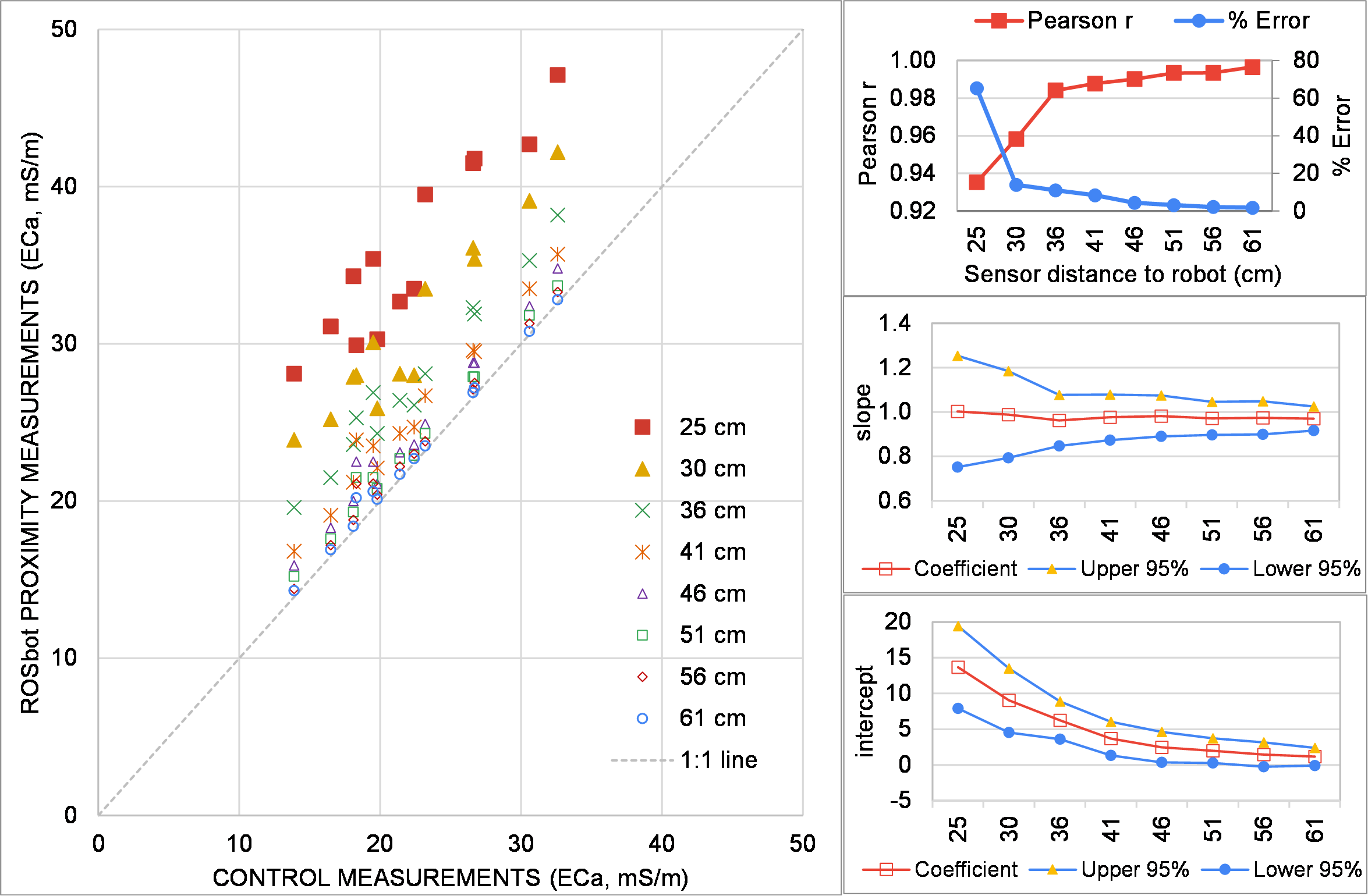}
    \vspace{-3pt}
    \caption{For the different field locations, ECa measurements were averaged for each distance $d_h$. The resulting average values were plotted against the 1:1 control line to determine the constant linear offset. Values that have a line with a slope of 1, parallel to the 1:1 line, and a Pearson correlation close to 1 can be treated as constant linear offsets. Distances greater than $46$\;cm represent the best candidates for linear offsets, while additional measurements are needed to determine the true relationship at closer distances.
    }
    \label{dh_plot}
    \vspace{-9pt}
\end{figure}

\textbf{Results:}
Our testing reveals that the (ideal) horizontal distance $d_h$ for the sensor would need to be at least $457$\;mm away from the main robot chassis in order to not to have a statistically significant influence on the EMI sensor measurements. At closer distances, the robot affects the sensor measurements, but it does not saturate the sensor. 
Table \ref{tab:prelim} contains the soil conductivity measurements at different test configurations. The column marked with $\infty$ denotes the control measurement at that location where the robot was not present. Figure~\ref{dh_plot} shows how the ECa measurements differ from the 1:1 control line along with the Pearson coefficient and regression slope and intercept for each distance. As the control vs ROSbot measurements all had a slope very close to 1, the influence of the ROSbot on the sensor measurements was considered constant in the range of conductivity measured in this experiment.

With reference to Table~\ref{tab:prelim}, we took measurements to determine the effect of the sensor angle $\theta$. Data suggest that the ideal orientation of the sensor would be $90^\circ$, that is the sensor should be mounted sideways and parallel to the direction of motion of the robot. However, this configuration would make the robotic platform unstable; hence, we elected to proceed with the horizontal sensor orientation (i.e., the sensor is mounted perpendicular to the direction of motion).

Mounting the sensor at the ideal distance from the robot would significantly impact its mobility, hence we elected to mount it a closer distance (as we describe in Section~\ref{subsec:mounting}) and identify a correction offset (as we describe in Section~\ref{subsec:discrete}) to account for the small, yet noticeable effect of the robot on the sensor's measurements. While the ideal placement would eliminate sensor bias, current best practices often mount the ECa sensors with metal brackets to farm equipment~\cite{willness_2021}, which introduces a measurement bias. Though further refinement of the robotic system could reduce bias, our platform conforms to standard practices.

Finally, in the field, the measured values did not noticeably change within the resolution of the measurements when the sensor was placed on top of a short spacer ($6.35$\;mm) or a taller one ($80$\;mm). To provide a large margin for the departure angle (as we describe in Section~\ref{subsec:mounting}), we elected to use $d_v = 50$\;mm which will provide sufficient ground clearance without obstructing the LiDAR's field of view.

\subsection{Vehicle Integration \& Traversal Tests}\label{subsec:mounting}

\textbf{Objective:}
Determine the maximum practical sensor mounting distance $d_h$ and angles $\alpha$ and $\beta$ that allow the robot to traverse uneven terrain with deviations of $\pm25$\;mm from level after mounting the sensor without tipping, stalling, or colliding the sensor into the ground.

\begin{figure}[!h]
\vspace{-4pt}
    \centering
    \includegraphics[trim={0cm 0.5cm 0cm 1cm},clip,width=0.95\linewidth]{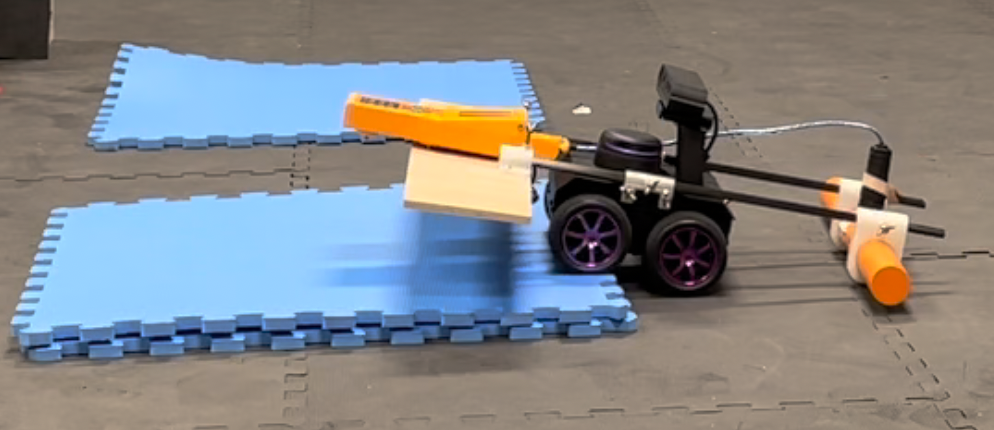}
    \vspace{-1pt}
    \caption{To test the system's maneuverability with the sensor attached, an obstacle course was created with several stacks of $12.5$\;mm thick foam tiles for a total height of $25$\;mm. The robot was driven over the tiles at various angles and speeds to ensure no unwanted portion of the system made contact with the ground.}
    \label{test_traversal}
    \vspace{-4pt}
\end{figure}

\textbf{Setup:}
After completing the integration of the sensor and its data logger (mounted at the back and front of the robot, respectively; see Fig.~\ref{test_traversal}), traversal tests were conducted in the lab using foam obstacles on a foam surface to confirm that the robot could indeed navigate the desired terrain (Fig.~\ref{test_traversal}). Sensor mounting components were designed so that the sensor distance $d_h$ can be adjusted during the traversal tests.

The sensor was mounted using a combination of carbon fiber rods and custom 3D printed brackets. The assembly using carbon fiber rods is an appropriate means to suspend the sensor behind the robotic platform since the rods have high strength-to-weight ratio, and are rigid and non-metallic. The lightweight nature of carbon fiber aids the design goal of maximum portability. Rigidity is beneficial because it minimizes the likelihood that the cantilever sensor arm will vibrate or flex while the robot is traversing obstacles. Finally, metal support rods were not an option since they would affect the conductivity measurements. Use of metal fasteners was minimized to reduce interference with the sensor, though they could not be completely avoided for the sensor mounting. 

\textbf{Results:}
While the ideal integration configuration would have no interference between the robotic platform and the sensor, the sensor placement is constrained by the need to have approach ($\alpha$) and departure ($\beta$) angles that allow the robot to traverse a field with deviations of $\pm25$\;mm from level terrain. As such, the ideal distance $d_h$ determined in Section~\ref{subsec:prelim} is not possible to achieve. Instead, the maximum distance that prevented tipping while still allowing the vehicle to traverse the desired terrain was empirically determined at $d_h=235$\;mm. 

Following the tests discussed in this and the previous section, and with reference to Fig.~\ref{fig:sketch}, the selected system mechanical parameters are summarized in Table~\ref{tab:params}.

\begin{table}[h]
\vspace{-1pt}
\caption{Implemented System Mechanical Parameters}
\label{tab:params}
\vspace{-3pt}
\begin{center}
\begin{tabular}{ c c c c c}
\toprule
$d_h$ & $d_v$ & $\theta$ & $\alpha$ & $\beta$\\
\midrule
$235$\;mm & $50$\;mm & $0^\circ$ & $18.4^\circ$ & $12.2^\circ$\\
\bottomrule
\end{tabular}
\end{center}
\vspace{-3pt}
\end{table}


\subsection{Comparative Roboticized- and Manually-collected ECa Measurement Tests}\label{subsec:discrete}

\textbf{Objective 1:}
Quantify the impact the mounting distance $d_h=235$\;mm has on the ECa measurements.

\textbf{Objective 2:}
Determine if the introduced measurement bias is linear or nonlinear.

\textbf{Setup:}
A row of olive trees was selected as a comparison region between roboticized- and manually-collected data. First, the robot with integrated sensor was driven down the row multiple times to collect ECa measurements along with GNSS coordinates. Then, the same measurements were collected by a human carrying the sensor. These paired measurements can be compared to determine if there are any transients from the robot during operation that could impact the EMI measurement. These tests also served to validate the robot's ability to traverse the field terrain, though these observations were qualitative not quantitative.



\begin{figure}[!h]
\vspace{-12pt}
    \centering
    \includegraphics[trim={0.8cm 0cm 1.2cm 0cm},clip,width=0.85\linewidth]{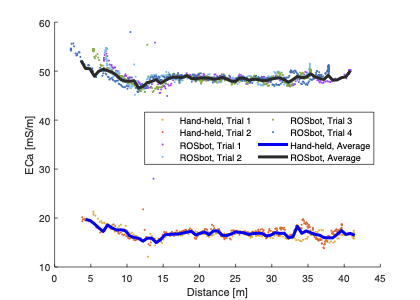}
    \vspace{-2pt}
    \caption{ECa Measurements Comparison. Raw data measured with hand-held approach (top) and ROSbot (bottom) are scattered in colors for different trials. Average measurements are plotted in lines.}
    \label{compare_man_robo}
    \vspace{-12pt}
\end{figure}

\textbf{Results:}
Since the mechanical constraints impacted the sensor placement, the goal here is to determine if the presence of the robot introduced a constant linear offset into the ECa Measurements. Figure~\ref{compare_man_robo} shows ECa measurements captured via a hand-held manual data acquisition process and when mounted on the robot. The roboticized- and manually-collected datasets were filtered to remove outliers, averaged, and then compared. The average of the hand-held ECa data set was $16.9$\;mS/m. The average of the dataset collected with ROSbot was $48.6$\;mS/m. The average difference was $31.7$\;mS/m. The slope of the linear relationship between the two datasets was $0.69$ with a Pearson correlation coefficient of $0.72$. A stronger correlation would be preferred, yet the obtained accuracy offers an acceptable trade-off between sensor measurement accuracy and compact robot design.

\subsection{Geospatial Soil ECa Tests}

\textbf{Objective:}
Generate ECa maps at a selected study site and compare the roboticized- and manually-collected measurement data. 

\textbf{Setup:}
In the most extensive field test conducted in this work, the study site was a  
$50$\;m $\times$ $30$\;m 
drip-irrigated olive orchard located within the Agricultural Experimental Station of the University of California in Riverside (33°58'24.5"N 117°19'10.3"W). Soils ($[0, 0.7]$\;m depth range) at the site were sandy loam, with sand content ranging between $54.6$\% and $65.9$\% \cite{BEAUDETTE20092119}. On March 12, 2021, two ECa surveys were carried out at the study site to compare hand-held measurements with these acquired with our proposed integrated robotic soil sensor platform. In addition to the ECa measurements, GNSS position was recorded to compare the results. The robot also recorded its pose (position and orientation), although these measurements were not available during the manual hand-held process. The hand-held survey collected 461 ECa measurements, whereas the ROSbot survey collected 6901. Collected data were sampled at the same frequency, but the robot was moving slower than the human counterpart, thus leading to the increased number of samples via the roboticized approach. 

\begin{figure}[!t]
    \vspace{6pt}
    \centering
    \includegraphics[width=0.995\linewidth]{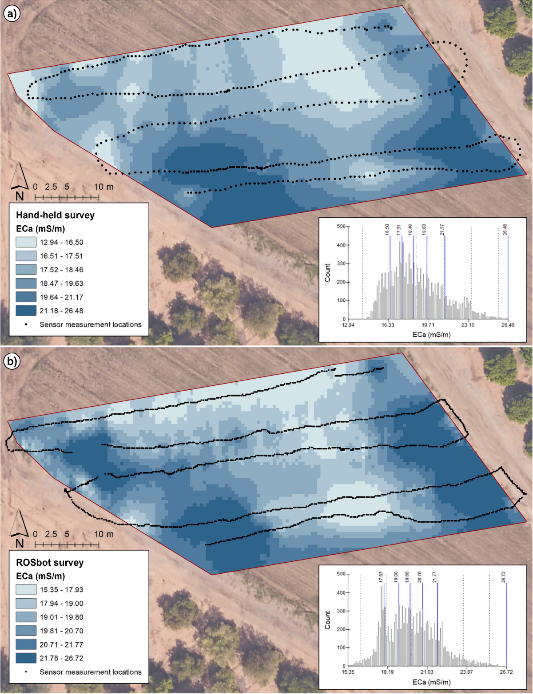}
    \caption{Maps of soil apparent electrical conductivity (ECa) for the 0-0.7m soil profile at the study site: a) hand-held survey and b) ROSbot semi-autonomous survey. Color scales are characterized with the quantile method. The maps' ECa frequency distribution are reported on a histogram on the bottom-right of each panel. }
    \label{spatial_map}
    \vspace{-9pt}
\end{figure}

Following commonly used data filtering procedures~\cite{CORDOBA201695,lesch2005apparent}, the ECa data were normalized with a natural logarithm transformation. Then, also in line with the standard of practice, any values outside the range of $\pm 2.5$ standard deviation around the average values were removed because deemed as outliers. The hand-held ECa dataset had $2$ outliers, whereas the ROSbot dataset had $137$. The increased number of outliers in roboticized measurements is linked to the fact that those measurements were substantially more compared to manually-collected ones. 
The average of the hand-held ECa dataset was $19.0$\;mS/m. The average ECa value measured with the ROSbot was $53.5$\;mS/m. The difference ($34.5$\;mS/m) was removed from all ECa measurements in the ROSbot dataset, to compare the variability and spatial distribution of the two datasets. In ArcMap 10.8 (ESRI, Redlands, CA), the ECa for the two datasets were spatially interpolated using simple kriging with exponential semivariogram to generate maps with $0.5$\;m $\times$ $0.5$\;m resolution. These spatial maps are displayed in Fig.~\ref{spatial_map}.

\textbf{Results:}
The ECa map obtained from the hand-held survey had mean = 18.78 mS/m, standard deviation = 2.31 mS/m, minimum = 12.94 mS/m, and maximum = 26.48 mS/m. The ECa map derived from the ROSbot survey had mean = 19.94 mS/m, standard deviation = 1.85 mS/m, minimum = 15.35 mS/m, and maximum = 26.72 mS/m (Table \ref{tab:mapstats}). The two maps revealed similar ECa spatial patterns, with the highest ECa values observed at the SE and SW portions of the study site, and the lowest ECa measured in the N and the center of the site. At the pixel-by-pixel level, the maps had a Pearson correlation coefficient = 0.65. This indicates that there are some inconsistencies between the maps, possibly associated with: non-constant influence of the ROSbot on the sensor measurements, different sensor distance to soil and tilt across the two surveys, and higher detail (and variance) captured in the ROSbot survey than in the hand-held one. Observed geolocation inaccuracies are because of the employed low-resolution GPS that also introduces inconsistencies.

While the robot collected data at a slower rate than manual operation, this actually provides more accurate measurements~\cite{SUDDUTH2001239}. Current commercial practices can suffer from sensor biases and inconsistent practices. A fleet of robots could provide more consistent data and broader coverage of field regions (such as near tree roots, under dense canopies) where a human could not regularly access.

\begin{table}[h]
\vspace{-3pt}
\caption{ECa Map Statistics. All values are in mS/m.}
\label{tab:mapstats}
\vspace{-6pt}
\begin{center}
\begin{tabular}{ l c c c c}
\toprule
Survey & $\mu$ & $\sigma$ & MIN & MAX\\
\midrule
Hand-held & 18.78 & 2.31 & 12.94 & 26.48\\
ROSbot & 19.94 & 1.85 & 15.35 & 26.72\\
\bottomrule
\end{tabular}
\end{center}
\vspace{-6pt}
\end{table}

\section{Discussion and Outlook}


\textbf{Contributions and Key Findings:}
Our robot prototype demonstrates the feasibility of conducting ECa surveys using an EMI sensor mounted on a small, portable robotic platform. The ECa geospatial measurements provide real-time spatial information that can be used to infer soil water status. Precision irrigation (i.e. for distinct zones in a field) or traditional irrigation (i.e. the whole field is managed uniformly) can be scheduled based on soil water status spatial information derived from the proposed platform. The size of the platform is particularly advantageous because it brings the ECa sensor closer to the sample regions of interest such as irrigation drip-lines and tree roots where current handheld, human operated surveys cannot access. Our integration approach highlights some of the design trade-offs with respect to the sensor position relative to the robotic platform, and quantifies how the robot might bias the sensor readings. Our system is capable of gathering the data necessary to create spatial maps and accurate spatio-temporal soil moisture information, which is key to increasing the environmental and economic sustainability of irrigation management in precision and traditional agricultural systems. 

\textbf{Directions for Future Work:}
Despite the promising initial capabilities, the system could be improved with a few modifications to 1) optimize the relationship between the sensor's signal to noise ratio and platform mobility, and 2) examine sensor performance with different soil types.
Additional ECa surveys should sample a wider variety of soil types not present in our initial data set. Current tested soils were sandy loam with sand content greater than 50\% and soil with greater variability in sand, clay, and silt content could produce different offsets. As part of these experiments, future robot iterations should test towing the sensor parallel to robot direction or mounting the sensor in different planes to minimize impact. 
Although the software foundations are in place with the sensor's ROS integration, the robot's software should implement more intelligent path planning and autonomous navigation capabilities. 
Finally, robot-gathered ECa data can pair with existing precision agriculture systems to determine optimal irrigation strategies.

\addtolength{\textheight}{-12cm}   




\section*{Acknowledgement}

The authors would like to thank the USDA-ARS U.S. Salinity Laboratory for granting the research team access to their fields to conduct experiments.



\bibliographystyle{IEEEtran}
\bibliography{agbot}

\end{document}